\documentclass[letterpaper]{article} % DO NOT CHANGE THIS
\usepackage{aaai23}  % DO NOT CHANGE THIS
\usepackage{times}  % DO NOT CHANGE THIS
\usepackage{helvet}  % DO NOT CHANGE THIS
\usepackage{courier}  % DO NOT CHANGE THIS
\usepackage[hyphens]{url}  % DO NOT CHANGE THIS
\usepackage{graphicx} % DO NOT CHANGE THIS
\usepackage{natbib}  % DO NOT CHANGE THIS AND DO NOT ADD ANY OPTIONS TO IT
\usepackage{caption} % DO NOT CHANGE THIS AND DO NOT ADD ANY OPTIONS TO IT
\usepackage{algorithm}
\usepackage{algorithmic}

\usepackage{multirow}
\usepackage{subfigure}
\usepackage{graphicx}
\usepackage[normalem]{ulem}
\useunder{\uline}{\ul}{}
\usepackage{newfloat}
\usepackage{listings}
\DeclareCaptionStyle{ruled}{labelfont=normalfont,labelsep=colon,strut=off} % DO NOT CHANGE THIS
\floatstyle{ruled}
\newfloat{listing}{tb}{lst}{}
\floatname{listing}{Listing}
\title{DC-Former: Diverse and Compact Transformer for Person Re-Identification}
\author{
    Wen Li\textsuperscript{\rm 1},
    Cheng Zou\textsuperscript{\rm 1},
    Meng Wang\textsuperscript{\rm 1},
    Furong Xu\textsuperscript{\rm 1},
    Jianan Zhao\textsuperscript{\rm 1}, \\
    Ruobing Zheng\textsuperscript{\rm 1},
    Yuan Cheng\textsuperscript{\rm 2}\thanks{Corresponding author},
    Wei Chu\textsuperscript{\rm 1}
}
\affiliations{
    \textsuperscript{\rm 1}Ant Group  \\
    \textsuperscript{\rm 2}Artificial Intelligence Innovation and Incubation (AI$^3$) Institute, Fudan University\\
{\{yinian.lw,wuyou.zc,darren.wm,booyoungxu.xfr,zhaojianan.zjn,zhengruobing.zrb,\\weichu.cw\}}@antgroup.com, cheng\_yuan@fudan.edu.cn

}

\usepackage{bibentry}
\begin{document}

\maketitle

\def\MethodName{DC-Former}  % Method Name

\begin{abstract}
In person re-identification (re-ID) task, it is still challenging to learn discriminative representation by deep learning, due to limited data. Generally speaking, the model will get better performance when increasing the amount of data. The addition of similar classes strengthens the ability of the classifier to identify similar identities, thereby improving the discrimination of representation. In this paper, we propose a Diverse and Compact Transformer (DC-Former) that can achieve a similar effect by splitting embedding space into multiple diverse and compact subspaces. Compact embedding subspace helps model learn more robust and discriminative embedding to identify similar classes. And the fusion of these diverse embeddings containing more fine-grained information can further improve the effect of re-ID. Specifically, multiple class tokens are used in vision transformer to represent multiple embedding spaces. Then, a self-diverse constraint (SDC) is applied to these spaces to push them away from each other, which makes each embedding space diverse and compact. Further, a dynamic weight controller (DWC) is further designed for balancing the relative importance among them during training. The experimental results of our method are promising, which surpass previous state-of-the-art methods on several commonly used person re-ID benchmarks. \textit{https://github.com/ant-research/Diverse-and-Compact-Transformer}

\end{abstract}

\section{Introduction}
Person re-identification (re-ID) aims at identifying person across different camera views, which is very important in many applications, such as intelligent surveillance, cross camera tracking and smart city. While re-ID has attracted great research interest and gained considerable development in recent years, there still exist some challenges~\cite{yan2021occluded, yang2021learning}, such as blur, low resolution, occlusion, illumination and viewpoint variation. These factors cause the intra-class distance of samples to be larger than the inter-class distance, which makes it challenging to retrieve pedestrians of correct identities.

\begin{figure}[t!]
\begin{center}
\includegraphics[width=8cm]{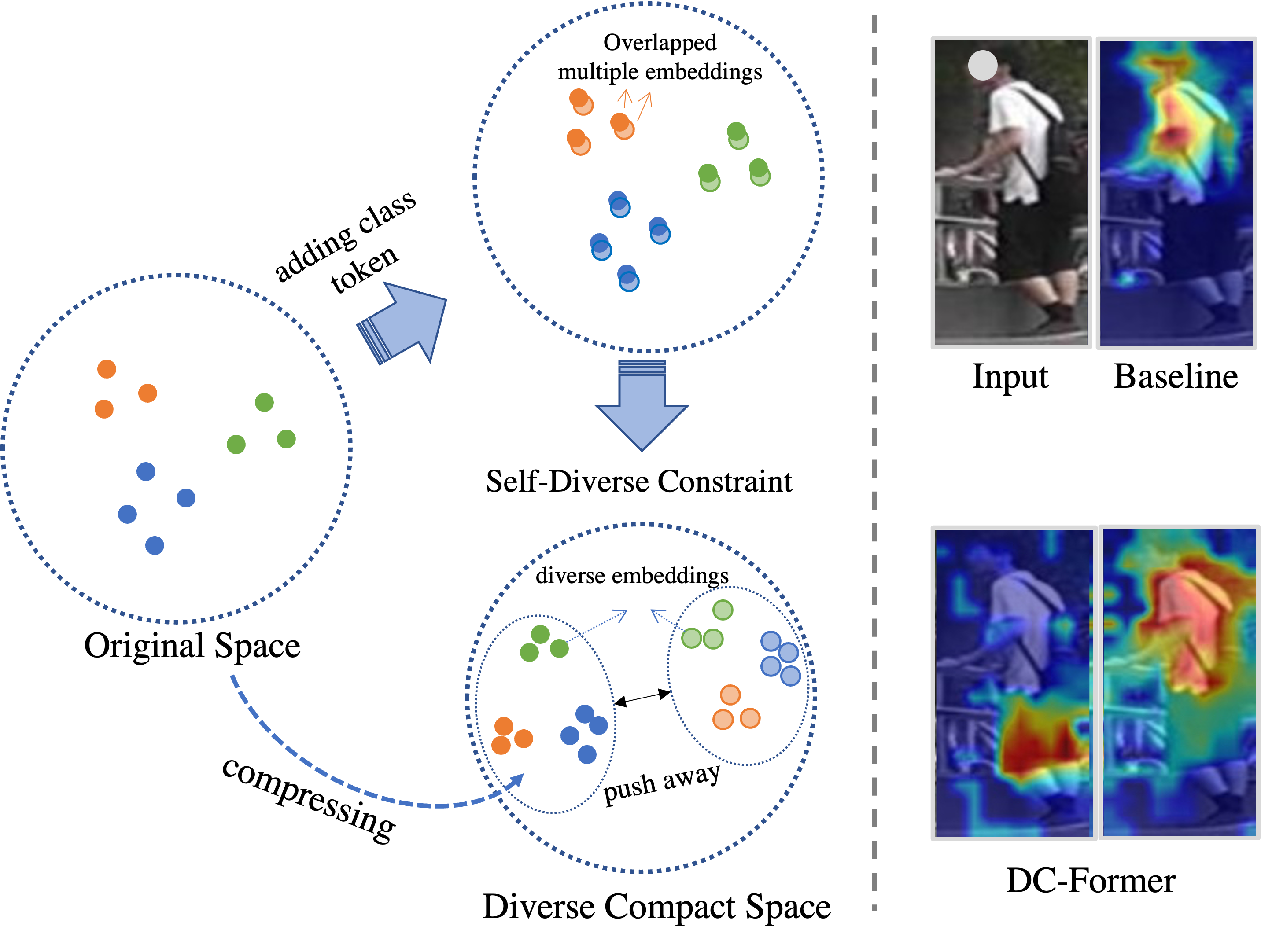}
\end{center}
\caption{Illustration of DC-Former for representation learning. On the left, each circle with dots denotes an embedding space. DC-Former uses multiple embeddings to represent each sample, and a self-diverse constraint is imposed on these embeddings to push them away. Finally, DC-Former obtains multiple diverse embedding subspaces for representation. And each subspace is more compact than the original space, which increases the identity density of embedding space to help model improve its discrimination for identifying similar classes. Figures on the right visualized by Grad-Cam~\cite{selvaraju2017grad} show that diverse embeddings from DC-Former focus on multiple different discriminative regions. And the fusion of them can provide more fine-grained information.}
\label{fig:inro}
\end{figure}

% Modified by wuyou
Plenty of efforts~\cite{transreid, AutoLoss_2022_CVPR, dpm2022lei} have been made recently to improve the performance of re-ID, and among which increasing the amount of training data may be the most powerful way. On the one hand, increasing more instances for each identity helps to recognize one person under different circumstances, extracting the most common and discriminative features for the same class, thus reducing intra-class distance. On the other hand, increasing more identities means that there is a higher chance to place more similar (easy-to-confuse) classes in the embedding space, which would help the model to extract discriminative features with larger inter-class distance for similar classes. Therefore, more identities and more instances for each identity, resulting in larger inter-class distance and smaller intra-class distance, make re-ID task easier.

As mentioned above, great potential lies in data. However, due to limited data, existing works concentrate more on other aspects, such as stronger backbone~\cite{swin}, metric loss~\cite{sun2020circle}, pre-trained model~\cite{luperson}, etc. As far as we know, few studies consider re-ID task from the perspective of data except for~\cite{grayscale,RandomErasing}, all of which are data augmentations, finally increasing the instances of each identity, but no study tries to increase the number of identity. Directly increasing the number of identity is impractical because it equals to adding more labeled data, which is expensive, but if there is a certain way that can simulate to increase the number of identity, it will also improve the performance of re-ID. Here we hypothesize that increasing the number of identity is equal to increasing the identity density in a given space. Reducing the size of the embedding space can make all the identities become more compact, so the relative amount of identity (identity density) increases. In a more compact embedding space, the reduction of inter-class distance makes it more difficult for model to distinguish samples from similar classes. Therefore, more discriminative and robust information is extracted to ensure that the classifier correctly identifies similar classes. 

In this paper, we propose a Diverse and Compact Transformer (DC-Former) that can achieve a similar effect of increasing identities of training data. As shown in Figure~\ref{fig:inro}, it compresses the embedding space by dividing the original space into multiple subspaces. More compact representations in the compressed embedding subspace helps model extract more discriminative feature to identify similar classes. And the embeddings of different subspaces are diverse, the fusion of them contains more information that can further improve performance. Specifically, multiple class tokens (CLSes) are used in vision transformer, and each CLS is supervised by an identity loss to obtain multiple representations. Then, a self-diverse constraint (SDC) is applied to CLSes to make the distribution of them as far as possible. In this way, the original space is divided into multiple subspaces. Due to the different learning status of different CLSes when dividing the space, some CLSes are pushed away while others are very close. A dynamic weight controller (DWC) is further designed for balancing the relative importance among them during training. Finally, each of compact subspaces learns a more robust representation. The experimental results of our method are promising, which surpass previous state-of-the-art methods on three commonly used person re-ID benchmarks.

The contributions of this paper are summarized as follows:
\begin{itemize}
\item We propose a DC-Former to get multiple diverse and compact embedding subspaces, each embedding of these compact subspaces is more robust and discriminative to identify similar classes. And the fusion of these diverse embeddings can further improve the effect of re-ID.
\item We propose a self-diverse constraint (SDC) to make embedding subspaces presented by each class token do not overlap. And a dynamic weight controller (DWC) is devised to balance the relative importance among multiple class tokens during training. 
\item Our method surpasses previous methods and sets state-of-the-art on three person re-ID benchmarks including \emph{MSMT17}, \emph{Market-1501} and \emph{CUHK03}.
\end{itemize}

\begin{figure*}[t]
\begin{center}
\includegraphics[scale=0.5]{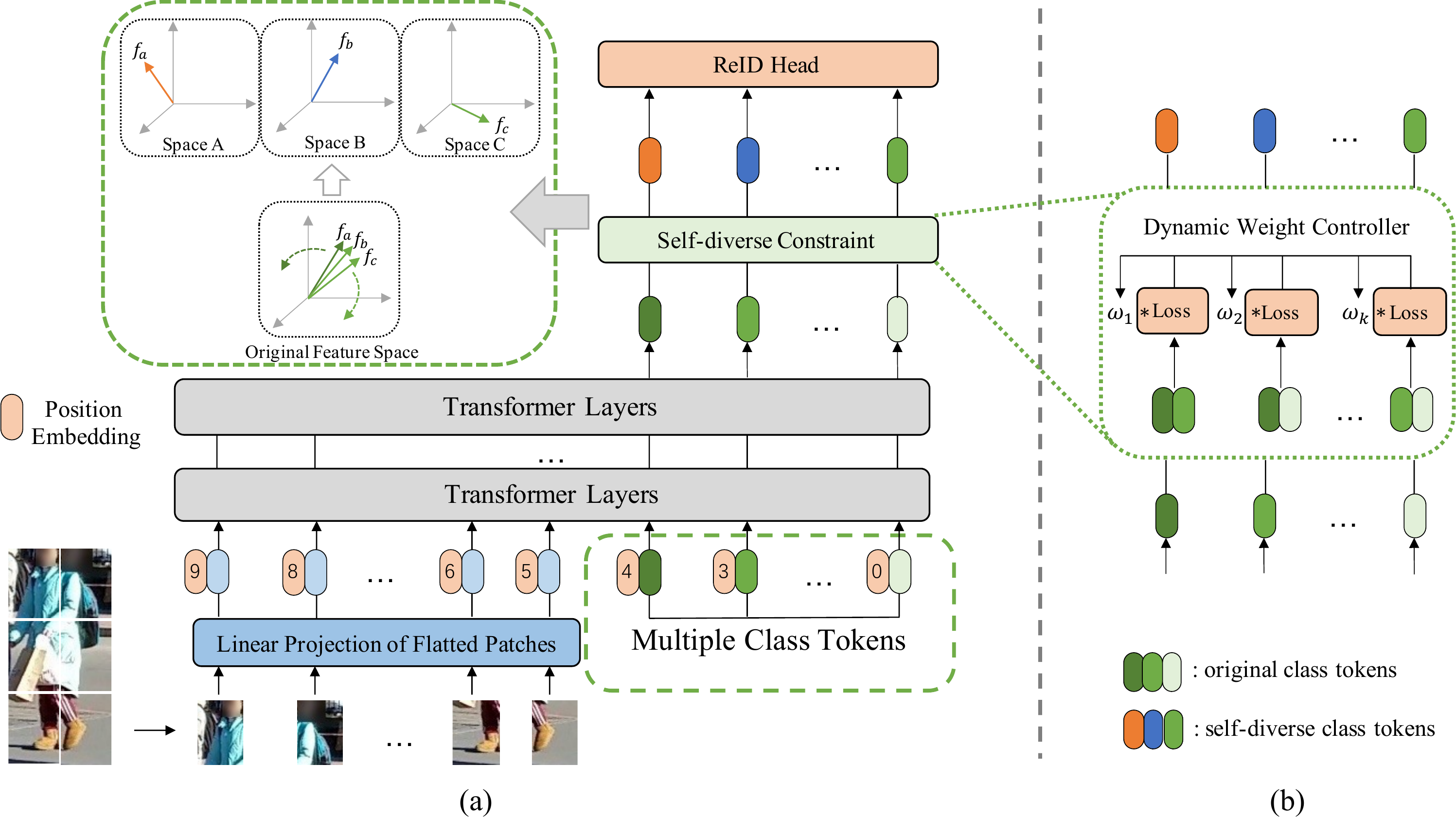}
\end{center}
\caption{The framework of \MethodName. Multiple class tokens concatenated with patch embeddings, adding positional embeddings, are fed into transformer encoder. Self-diverse constraint is employed on these class tokens in the last transformer layer to push them far way from each other, leading to diverse representation spaces. Then they are each supervised by a re-ID head, which contains a triplet loss and a classification loss. During training, a dynamic weight controller is used to dynamically adjust the constraint loss for easier optimization. $f_a$, $f_b$ and $f_c$ are different representations of the same object. }
\label{fig:arch}
\end{figure*}

\section{Related Work}
\subsection{Image-based re-ID}
Re-ID can be conducted on either images \cite{transreid} or videos \cite{zhao2021phd} . Recent image-based tasks mainly focus on person re-ID and vehicle re-ID.
The studies of person re-ID have paid attention to feature representation learning \cite{chen2019self,suh2018part} and deep metric learning \cite{zheng2017person,deng2018image}.
For the feature learning strategy, Typically, there are three main categories for the feature learning strategy, including global feature \cite{chen2019self}, local feature \cite{suh2018part} and auxiliary feature (e.g., domain information \cite{lin2017consistent}, semantic attributes \cite{lin2019improving}, viewpoint information \cite{zhu2020aware}).
As for the deep metric learning, many existing works have  designed loss functions to guide the feature representation learning, which can be divided into three groups, i.e., identity loss \cite{zheng2017person}, verification loss \cite{deng2018image} and triplet loss \cite{hermans2017defense}.

\subsection{Representation Learning in re-ID} 
Re-ID is one of fine-grained task which has far intra-class distance and close inter-class distance. Extracting discriminative features to distinguish similar classes is challenging. To minimize intra-class distance and maximize inter-class distance, proxy-based loss \cite{sce,imagenet} and pair-based loss \cite{triplet,hermans2017defense} are commonly used to push the different class away and pull the same class close. Some works use attention-base method \cite{Huynh_2020_CVPR} to discover discriminative features or enhance low-discriminative features, which helps disciminative feature representation. For example, DAM \cite{dam} iteratively identifies insufficiently trained elements and improves them. And some works \cite{RandomErasing,srivastava2014dropout} impose some regulation operations (e.g., random erasing and dropout) to prevent overfitting. Self-supervised representation learning \cite{he2019moco,chen2020mocov2,dino,CCL} use contrastive loss to maximize the agreement of features from same image with different augmentations.

\subsection{Transformer-based re-ID}
Before vision transformer, CNN-based methods have achieved absolute advantages on re-ID task. Methods like PCB~\cite{pcb}, MGN~\cite{mgn}, etc., partition an image into several stripes to obtain local feature representations for each stripe with multiple granularities. With the success of transformer in the field of computer version, it's also widely used in person re-ID. Compared with CNN-based method, ViT can keep more detailed information because these is no convolution and dowmsampling operators, which is more suited for fine-grained task like re-ID. TransReID~\cite{transreid} is the first pure transformer-based method on re-ID, it proposes a jigsaw patches module (JPM) which shuffles patch embeddings and re-groups them for further feature learning to extract several local features and aggregates them to get robust feature with global context. TransReID-SSL \cite{transreid-ssl} uses a massive person re-ID dataset $LUPerson$ \cite{luperson} to train a stronger pre-trained model by DINO \cite{dino}.

Many works~\cite{zhu2021aaformer,PAT,oh-former,TPM} devote to extract partial region representation by transformer. For example, AAformer \cite{zhu2021aaformer} uses the additional learnable vectors of `part tokens' to learn the part representations by clustering the patch embeddings into several groups and integrates the part into the self-attention for alignment. Some methods \cite{PAT,TPM} use transformer to learn several different part-aware masks to get partial feature. And some works aim to fuse features at different granularities. For example, HAT \cite{HAT} put hierarchical features at different granularities from CNN backbone into transformer to aggregate them. These methods have one thing in common that multiple different features are extracted and integrated to obtain more robust representation. However, they impose constraints on specific goals like focusing only on local receptive fields or different granularity which are limited to extract representation that the model really needs. It is insightful to learn multiple feature space by model itself through certain implicit constraint.

\section{Methodology}

\subsection{Overview}
For enhancing the identity density to increase the ability of discriminating similar classes, the original embedding space is divided into multiple diverse and compact subspaces. The proposed method consists of three parts: a ViT-based network with multiple class tokens to produce multiple embedding spaces, a self-diverse constraint loss (SDC) to push these embedding spaces far away from each other, and a dynamic weight controller (DWC) to balance the relative importance among class tokens during training as the token number increases.

\subsection{Network Architecture}
\label{sec:3.1}
% \textbf{Multiple Class Tokens.} 
To construct multiple different embedding spaces for one single input image, an architecture with multiple parallel output features is required. Most multi-feature representation methods usually extract different features from different granularities or different partial regions by multiple branches \cite{mgn,pcb}. The multiple features generated by these branches are often individual and have no interactions with each other during train, which may cause multiple embedding spaces overlapping and homogenous.

In vision transformer, thanks to stacked self-attention modules, information flows from patch embeddings to one class token, layer by layer gradually and autonomously. The class token acts as an information collector here, it receives information from each patch and outputs the summary according to its prior knowledge (learnable parameters). It's natural to come up with an idea that if there exist multiple class tokens acting as multiple information collectors, there will be a chance to gather different information from patch embeddings. Then, multiple different embedding spaces can be supported by multiple class tokens. Therefore, ViT with additional class tokens is chosen to be the main achitecture.

Figure~\ref{fig:arch} illustrates the proposed framework. The input image is divided into $H \times W$ patches, and then each patch is mapped to a vector of dimension $D$ through a trainable linear projection. The output of this projection is denoted as patch embeddings $P^{C \times D}$, where $C$ is the number of patches. Then multiple learnable embeddings  of dimension $D$ are concatenated as a sequence, denoted as class tokens $f^{N\times D}$, where $N$ is the number of class tokens. After that, class tokens and patch embeddings are concatenated to get a vector sequence $X^{(N + C) \times D}=[f^{N\times D}; P^{C \times D}]$. Next, this sequence added with a learnable positional embedding sequence is fed into transformer layers to get multiple representations. 

Notably in our design, multiple class tokens are mutually visible in self-attention layers, which helps the model to converge because of sharing intermediate features. For one certain class token, it receives information not only from all the patch embeddings but also from other class tokens, which improves the information acquisition efficiency.

\begin{figure}[]
\begin{center}
\subfigure[w/o SDC]{
\includegraphics[width=4cm]{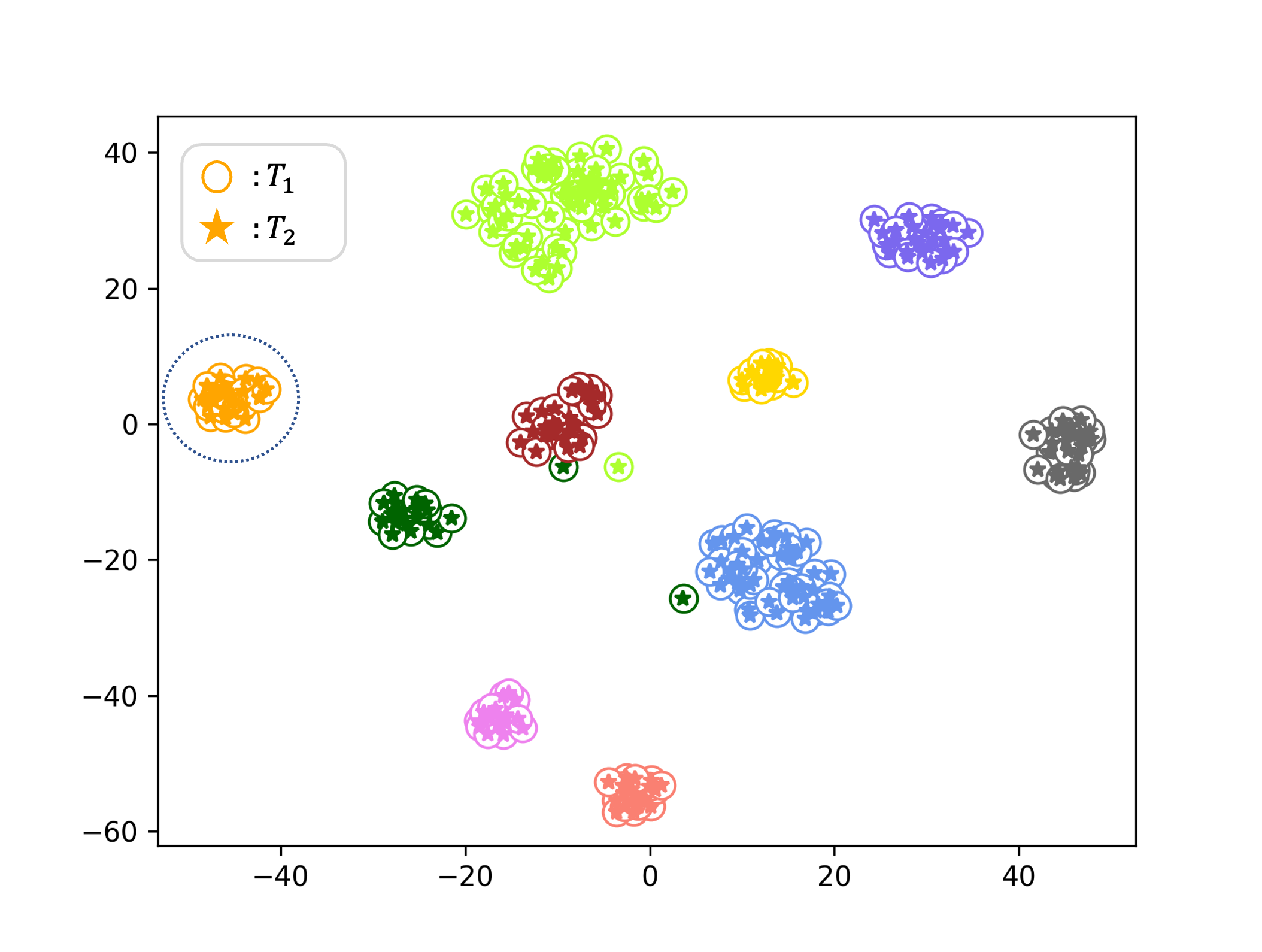}
}
\hspace{-0.4cm}
\subfigure[with SDC]{
\includegraphics[width=4cm]{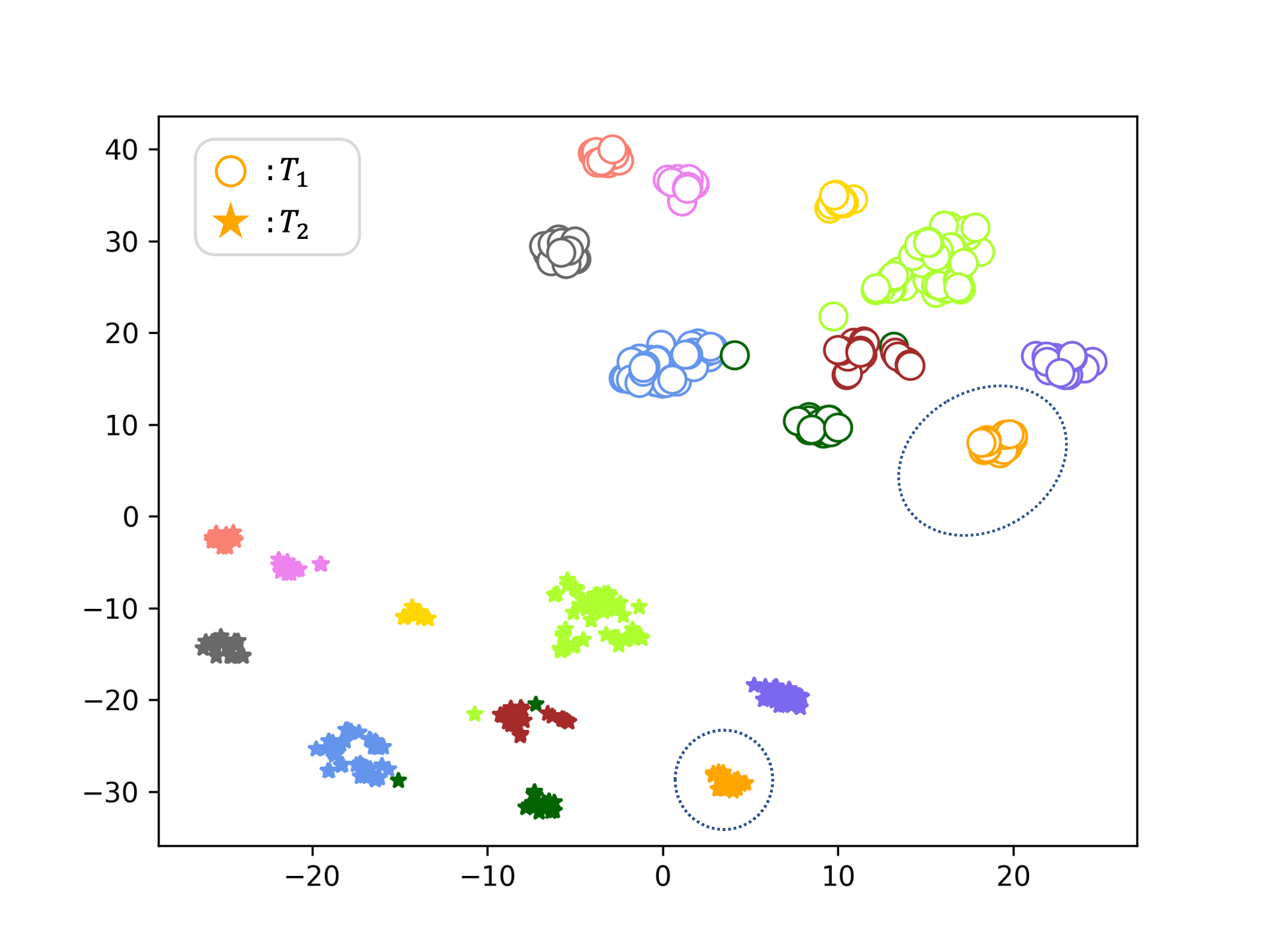}
}
\caption{Feature distributions visualized by t-SNE of two class tokens. Different colors represent different classes, marker $\star$ stands for feature points from one class token and $\bigcirc$ from the other. The data points are sampled from \emph{MSMT17} test set. (a) Without SDC, the feature points from two class tokens are heavily overlapped, which implies that the embedding spaces of these two class tokens are very close. (b) With SDC, the feature points from two class tokens can be easily separated into two parts without any overlap, and each embedding space become more compact.}
\label{fig:feat_space}
\end{center}
\end{figure}

\subsection{Self-diverse Constraint Loss}
\label{sce:diverse_loss}
If there are multiple different embeddings to represent each samples, the embedding space will become more compact, which helps model improve embedding's discrimination of similar classes. With multiple class tokens, the proposed framework can output multiple embeddings in parallel. But in practice, the learned embedding spaces overlap with each other because these class tokens are homogeneous in structure,  their learning goals are the same, and there is no external energy to push them to be different. 

One way to learn different embedding spaces is to change the learning objectives, DeiT~\cite{deit} proposed a distillation token to learn the output logits of another CNN model, which requires training two models and much manual efforts, making the training procedure complex. Considering that class token is computed as a weighted sum of $V$, although multiple class tokens share the same K-V pairs from patch embeddings $P^{C \times D}$, they can still get different attentions from $P^{C \times D}$ if the query $Q$ is different. In other words, class tokens can learn different information as long as they are different. 

Here we humbly hypothesize that the more difference between class tokens, the farther the distance between their embedding space, which pushes overlapping embedding spaces away from each other. Not only do we want the class tokens to be different from each other, but we want to maximize the difference between them. In the metric of cosine similarity, if two vectors are orthogonal, i.e., $cos(*,*)=0$, they are irrelevant, and the distance between them are extremely large. If the class tokens are orthogonal to each other, each embedding space of them is compressed more tightly in the finite space to ensure their distance is maximized.

We propose a self-diverse constraint loss (SDC) to constrain the relationship between class tokens.
It forces the class tokens to be orthogonal to each other and can be expressed as:
\begin{equation}
\label{loss:sdc}
    L_{sdc} =  \frac{1}{C_{N}^{2}}{\textstyle{\sum_{i}^{}} }^{} {\textstyle \sum_{j}^{}}\nu _{ij}, \ i<j, \ \ i,j=1,...,N
\end{equation}
where $\nu _{ij} = \left | cos(f_i, f_j) \right |$, and  $f_i$, $f_j$ indicates any two class tokens of  $f^{N\times D}$. Self-diverse constraint loss is employed on these class tokens to make sure that these embedding subspaces is far apart from each other. It makes each embedding subspace compact and helps model extract more discriminative feature to identify similar classes. 

The representations from different embedding subspaces contain not only identity information, but also features from different perspectives, i.e., coarse/fine-grained granularity, global/local region, and other unrecognized complementary aspects. The fusion of them can further facilitate robust and perturbation-invariant feature representation for re-ID tasks.

\begin{table*}[t]
% \scriptsize
\caption{Comparisons with state-of-the-art methods on person re-ID benchmarks. ${368\uparrow}$ and ${384\uparrow}$ denote the input images are resized to $368 \times 128$ and $384 \times 128$, otherwise $256 \times 128$. Best results for previous methods are underlined and best of our methods are labeled in bold.}
\label{tab:sota_person}
\vspace{-0.2cm}
\centering
\begin{tabular}{l|c|cc|cc|cc|cc}

\hline
\multicolumn{1}{c|}{\multirow{2}{*}{Methods}} & 
\multicolumn{1}{c|}{\multirow{2}{*}{Publications}} & 
\multicolumn{2}{c|}{\emph{MSMT17}} & 
\multicolumn{2}{c|}{\emph{Market-1501}} & 
\multicolumn{2}{c|}{\emph{CUHK03-l}} & 
\multicolumn{2}{c}{\emph{CUHK03-d}}\\
\multicolumn{1}{c|}{} & \multicolumn{1}{c|}{} & mAP & R1 & mAP & R1 & mAP & R1 & mAP & R1 \\
\hline
% GLDFA-Net~\cite{GLDFA-net} & 58.7 & 79.8 & 86.1 & 94.8 & 68.9 & 73.1 & 68.4 & 72.3 \\
PFD~\cite{wang2022pfd} & AAAI & 64.4 & 83.8 & 89.7 & 95.5 & - & -  & - & - \\
TransReID~\cite{transreid} & ICCV & 67.4 & 85.3 & 88.9 & 95.2 & - & - & - & - \\
DPM~\cite{dpm2022lei} & ACM'MM & - & - & 89.7 & 95.5 & - & - & - & - \\
GASM$^{384\uparrow}$~\cite{GASM} & ECCV & 52.5 & 79.5 & 84.7 & 95.3 & - & - & - & - \\
CDNet$^{384\uparrow}$~\cite{CDNet_2021_CVPR} & CVPR & 54.7 & 78.9 & 86.0 & 95.1 & - & - & - & - \\
AutoLoss$^{384\uparrow}$~\cite{AutoLoss_2022_CVPR} & CVPR & 63.0 & 83.7 & {\ul 90.1} & {\ul 96.2} & 74.3 & 75.6 & - & -\\
AAformer$^{384\uparrow}$~\cite{zhu2021aaformer} & - & 63.2 & 83.6 & 87.7 & 95.4 & {\ul 77.8} & {\ul 79.9} & {\ul 74.8} & {\ul 77.6}\\
OH-former$^{368\uparrow}$~\cite{oh-former} & - & 69.2 & {\ul 86.6} & 88.7 & 95.0 & - & - & - & - \\
TransReID$^{384\uparrow}$~\cite{transreid} & ICCV & {\ul 69.4} & 86.2 & 89.5 & 95.2 & - & -  & - & - \\
\hline
\MethodName & - & 69.8 & 86.2 & 90.4 & 96.0 & 79.4 & 81.6 & 77.5 & \textbf{80.1} \\
\MethodName$^{384\uparrow}$ & - & \textbf{70.7} & \textbf{86.9} & \textbf{90.6} & 96.0 & \textbf{83.3} &  \textbf{84.4} & \textbf{77.5} & 79.6 \\
\hline
\end{tabular}
\end{table*}

\subsection{Dynamic Weight Controller}
\label{sec:3.2}

To obtain more compact and robust features, more class tokens are needed to get more different embedding subspaces. However, as the number of class tokens increases, it becomes harder to optimize because there are more pairs of class tokens, each of which is required to be orthogonal.

In experiments, we find that when the number of class tokens increases to a certain number, the loss in Eq.~\ref{loss:sdc} cannot be minimized as expected.  Although some of the pairs have low cosine similarity, others are still very 
similar (cosine similarity close to 1). Actually it only learns less embedding space than we expect, and those class token pairs with higher self-diverse constraint loss are not optimized well in training. The reason for this is that randomness in the training process makes it easier for some pairs to be pushed farther apart while others harder. Therefore, it's necessary to change the relative importance among class token pairs for self-diverse constraint loss while training.

We propose a dynamic weight controller (DWC) to dynamically adjust the loss weight of each pair during training on the fly. Instead of simply averaging these pair losses as in Eq.~\ref{loss:sdc}, the loss of each pair is re-weighted by its own softmax-normalized loss. The weight of each loss can be defined as:

\begin{equation}
    \omega _{ij} = \frac{exp(\nu _{ij})}{{\textstyle{\sum_{m}^{}} }^{} {\textstyle \sum_{n}^{}} exp(\nu _{mn})}, \\ m<n, \ \ m,n=1,...,N
\end{equation}

So, the balanced self-diverse constraint loss is defined as:

\begin{equation}
    L_{SDC} =  {\textstyle{\sum_{i}^{}} }^{} {\textstyle \sum_{j}^{}}\omega _{ij}\nu _{ij}, \ i<j, \ \ i,j=1,...,N
\end{equation}

Those pairs with smaller cosine similarities are given smaller weights, while larger ones are given larger weights to make the model focus more on similar pairs. In this way, pairs can be learned more evenly so that they can be all orthogonal to each other.

\subsection{Objective Function}
\label{sec:3.3}
In training, in order to ensure that each class token has the ability to distinguish identities, they are each supervised by cross-entropy loss for classification (ID loss) after normalized by BNNeck~\cite{Luo_2019_CVPR_Workshops}. To pull the samples of the same class closer and push the samples of different classes far away, the triplet loss with soft-margin is used to mine hard example in each embedding subspace and can be calculated as:
\begin{equation}
 L_{triplet} = log[1+exp(|| f_a - f_p ||_2^2 - || f_a - f_n ||_2^2 )]
  \label{eq:1}
\end{equation}
The overall objective function is:
\begin{equation}
 L_{total} = \frac{1}{N}  {\textstyle \sum_{i=1}^{N}} (L_{ID}^{i} + L_{triplet}^{i}) + \lambda L_{SDC}
  \label{eq:2}
\end{equation}
During inference,  all the class tokens are concatenated to represent an image.

\section{Experiments}

\subsection{Experimental settings}

\textbf{Implementation Details.}
 We apply VIT-B/16~\cite{vit} as our backbone, it contains 12 transformer layers with hidden size of 768 dimensions. Overlapping patch embedding (step size = 12) and SIE~\cite{transreid} are also used in our experiments. All the images are resized to $256 \times 128$ unless other specified. The training images are augmented with random horizontal flipping, padding, random cropping, random erasing~\cite{RandomErasing}, and random grayscale~\cite{grayscale}. The initial weights of the models are pre-trained on ImageNet.

The batch size is set to 64 with 4 images per ID. SGD optimizer is employed with a momentum of 0.9 and a weight decay of 1e-4. The learning rate is initialized as 0.032 with cosine learning rate decay. All the experiments are performed on 4 Nvidia Tesla V100 GPUs. 

\textbf{Datasets and Evaluations.}
The proposed method is evaluated on four widely used person re-ID benchmarks, i.e., \emph{MSMT17} \cite{msmt}, \emph{Market-1501} \cite{market} and \emph{CUHK03} \cite{cuhk}. Mean Average Precision (mAP) and Cumulative Matching Curve (CMC) are used to evaluate the performance of re-ID tasks.

\begin{figure}[!t]
\begin{center}
\includegraphics[width=8cm]{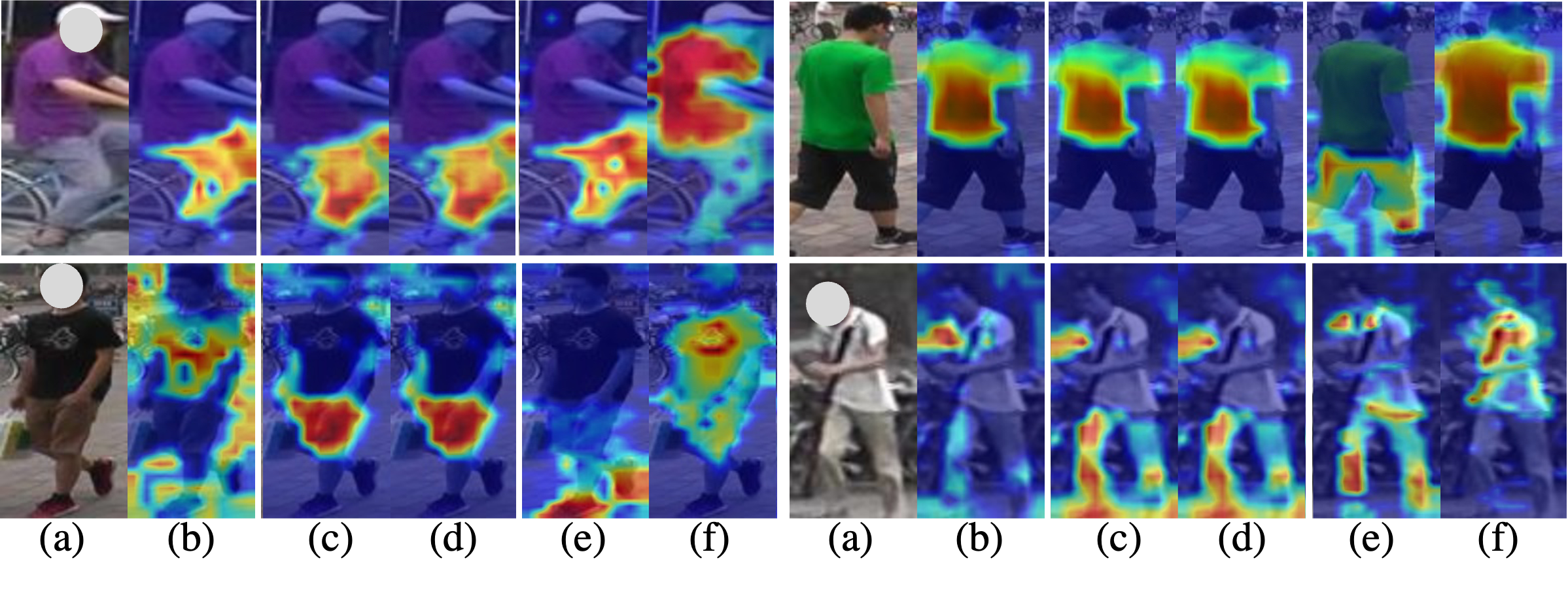}
\caption{Grad-Cam visualization of attention map on \emph{Market-1501}. (a) Input image. (b) Baseline. (c)-(d) Two class tokens without SDC. (e)-(f) Two class tokens with SDC. As can be seen, self-diverse constraint loss makes multiple class tokens to focus on different discriminative regions.}
\label{fig:grad_cam_2tokens}
\end{center}
\end{figure}

\subsection{Comparisons with State-of-the-Arts}
To verify the effectiveness of the proposed method, experiments are conducted on three commonly used person re-ID benchmarks in Table~\ref{tab:sota_person}. 
On \emph{MSMT17}, our method outperforms previous SOTA methods (e.g., TransReID) by a large margin especially on mAP (+2.4\%) with $256\times 128$ resolution, and also achieves the best performance with higher resolution $384\times 128$. 
On \emph{Market-1501}, our method achieves the best performance on mAP and comparable performance on Rank-1.
On \emph{CUHK03}, our method achieves absolute superiority which outperforms previous SOTA method (AAformer) by a large margin both on mAP (+4.2\%) and Rank-1 (+2.5\%).

\begin{table}[]
\caption{The ablation study of SDC on \emph{MSMT17}. $T_1$ and $T_2$ denote two class tokens, $Cat.$ denotes the concatenation of the two class tokens.}
\label{tab:SDC_person}
\centering
\begin{tabular}{ccccccc}
\hline
\multirow{2}{*}{} & \multicolumn{3}{c}{mAP} & \multicolumn{3}{c}{Rank-1} \\
                         & $T_1$   & $T_2$  & $Cat.$  & $T_1$    & $T_2$   & $Cat.$   \\
\hline
Baseline                 & -      & -     & 66.1   & -       & -      & 84.6    \\
w/o SDC                  & 66.9   & 66.9  & 66.9   & 84.8    & 84.8   & 84.8    \\
with SDC                 & 67.9   & 68.1  & \textbf{68.3}   & 85.3    & 85.4   & \textbf{85.5}   \\
\hline
\end{tabular}
\end{table}

\subsection{Ablation Study}
% \textbf{SDC Evaluation.} 
Experiments are conducted to study the effectiveness of self-diverse constraint loss (SDC), intra-class and inter-class distance of compressed embedding space, the hyper-parameters $\lambda$ and token number $N$, the effectiveness of dynamic weight controller (DWC), and the effectiveness of training with a smaller amount of identity.

\textbf{Impact of Multiple Class Tokens and SDC.} 
The effectiveness of multiple class tokens is validated on \emph{MSMT17} in Table~\ref{tab:SDC_person}. The baseline has only one class token. Adding one additional class token to the baseline (totally two class tokens) provides +0.8\% mAP improvement on \emph{MSMT17}. During training, each class token aggregates information not only from patch embeddings but also from other class tokens, which improves the information acquisition efficiency. But in further study, we find that the similarity between these two class tokens is very high, i.e., 0.999, which implies that there is no difference between them, so there is no improvement after fusion. And continuing to increase the number of tokens cannot continue to improve the performance.

With SDC imposed on these two class tokens, the cosine similarity of them becomes 0.007, which is very close to 0.0, implying that these two class tokens have learned two non-overlapping representation subspaces. Figure~\ref{fig:feat_space} illustrates the feature distributions of two class tokens, and they are separated to each other. Also, the performance of these two class tokens are both improved by a large margin especially on mAP, more than +1.0\% improvement, which means both tokens have learned more robust features. After concatenation, feature fusion further improves the performance, reaching 68.3\% mAP and 85.5\% Rank-1, which is +2.2\% mAP and +0.9\% Rank-1 higher than the baseline. The attention map visualized in Figure~\ref{fig:grad_cam_2tokens} shows that both two tokens correctly represent the foreground part of the object. Moreover, the two tokens represent different embedding spaces, and the information they represent is also different. Compared with the baseline, two class tokens with SDC has captured more fine-grained features. The fusion of multiple class tokens helps the model to learn more discriminative representation.

\begin{figure}[]
\centering
\subfigure[Baseline]{
\vspace{-0.2cm}
\includegraphics[width=4cm]{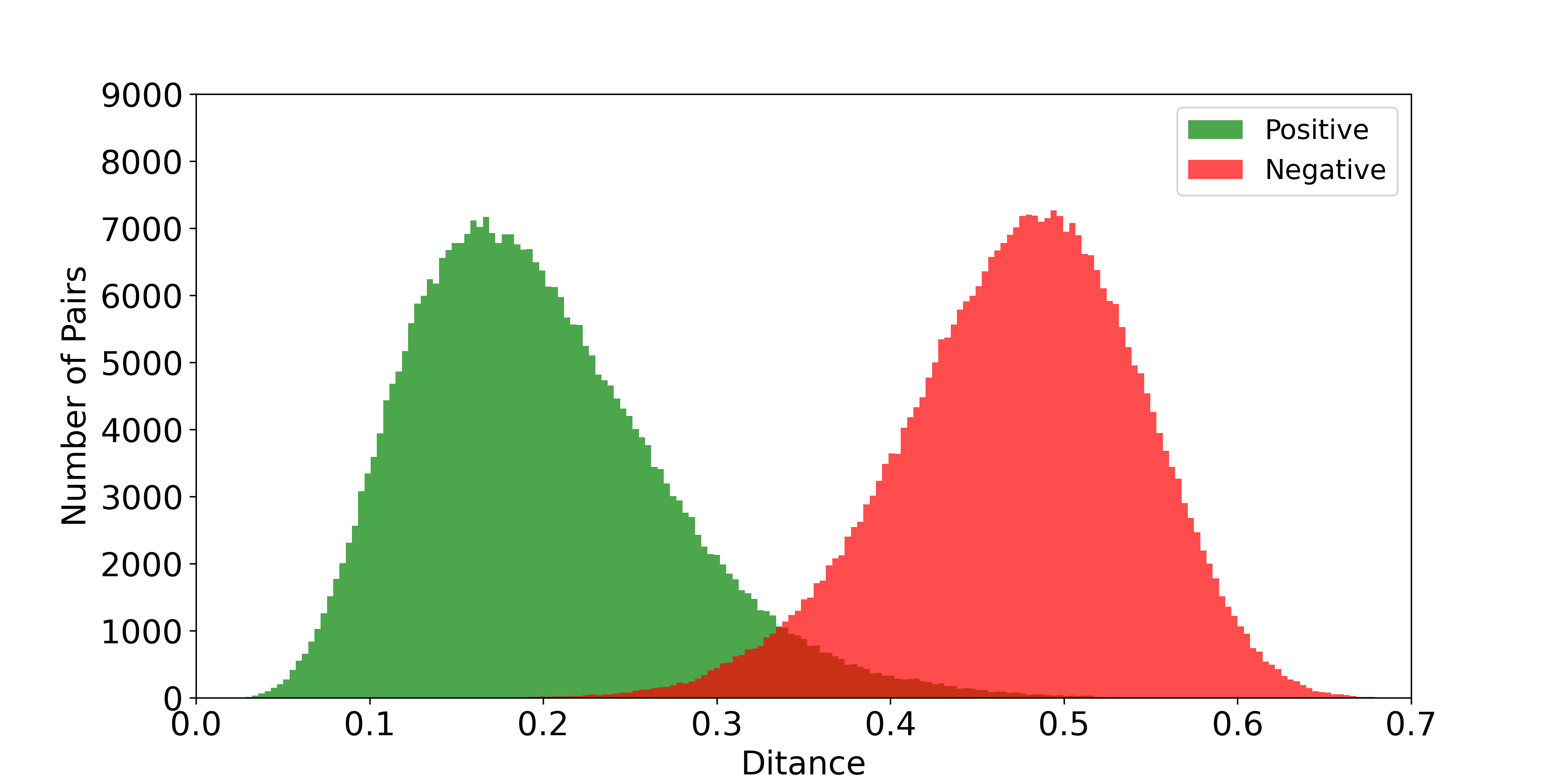}
}
\vspace{-0.2cm}
\hspace{-0.4cm}
\subfigure[$Cat(T_1,T_2)$]{
\includegraphics[width=4cm]{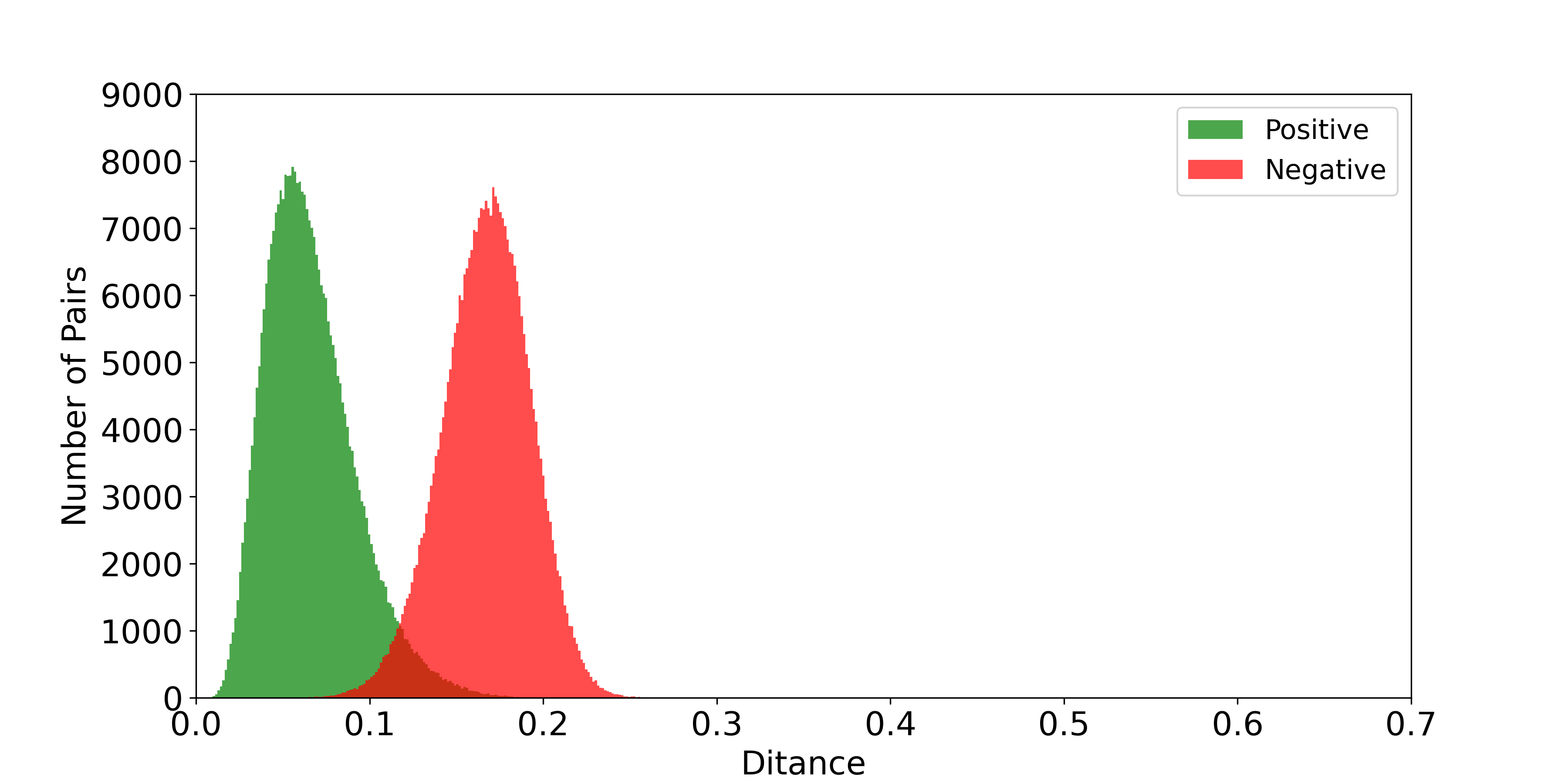}
}
\subfigure[$T_1$]{
\includegraphics[width=4cm]{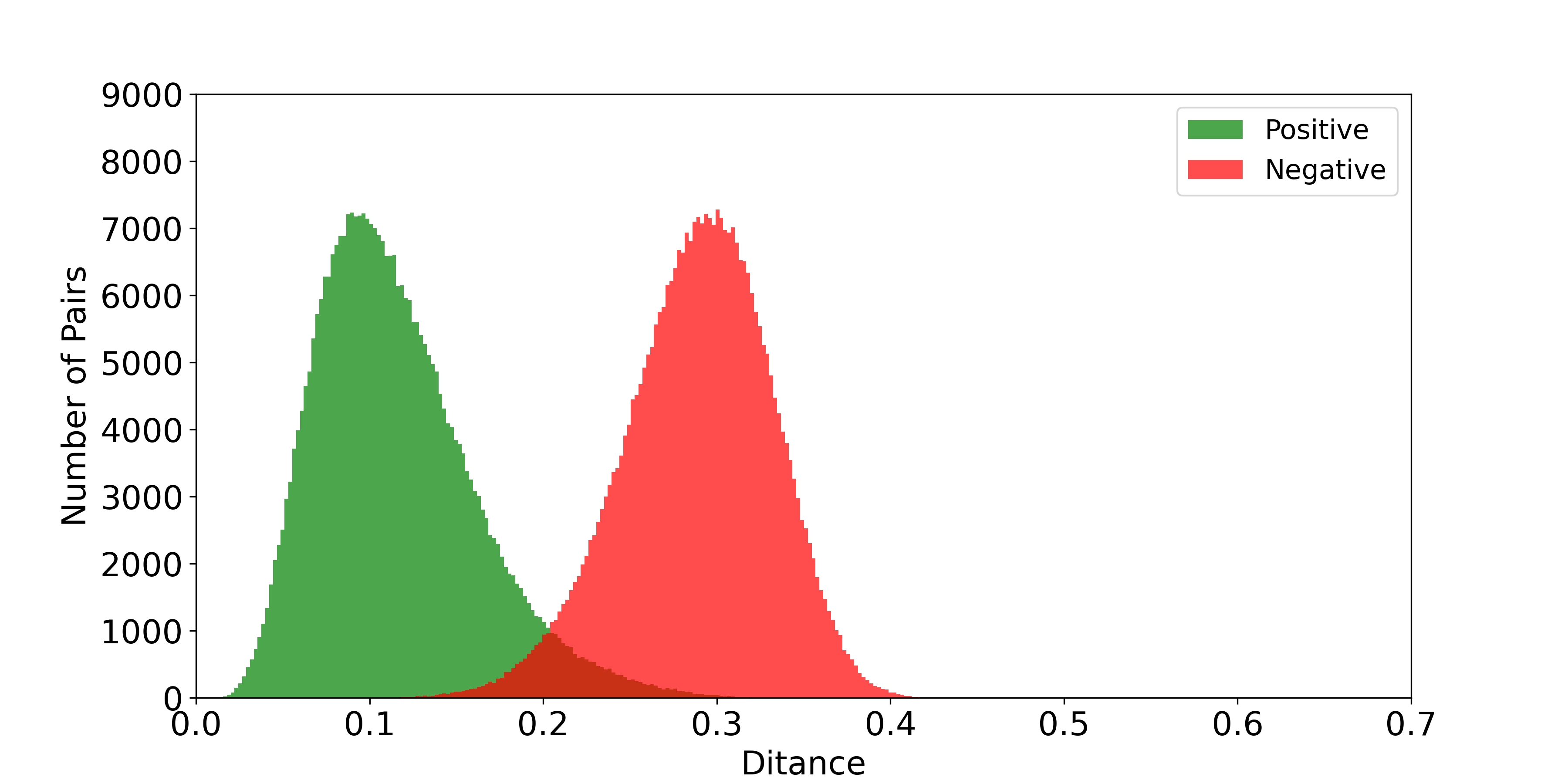}
}
\hspace{-0.4cm}
\subfigure[$T_2$]{
\includegraphics[width=4cm]{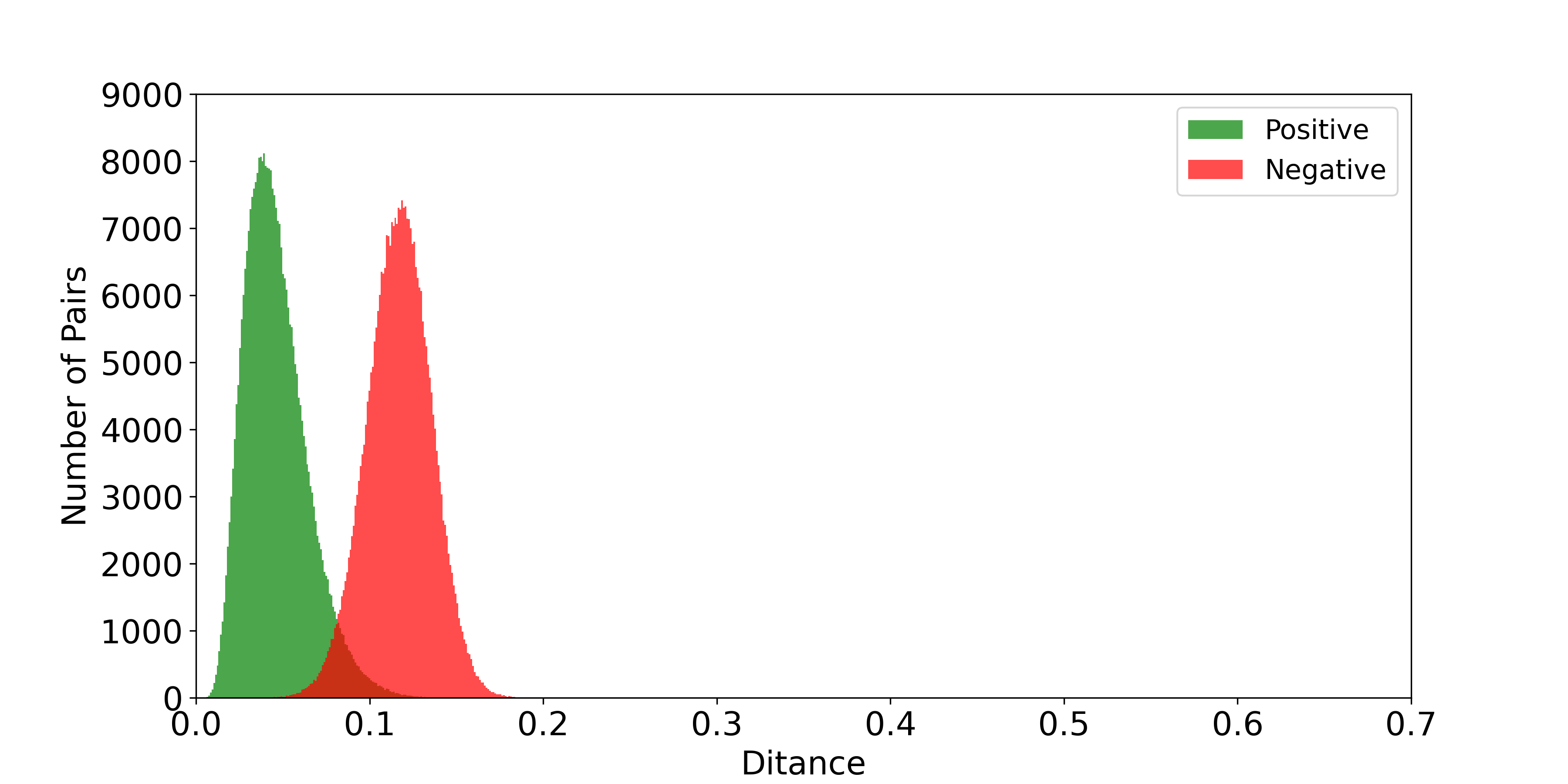}
}
\caption{The distance of positive and negative pairs in \emph{MSMT17} test set. The positive/negative pair denotes that two sample from $Query$ and $Gallery$ are the same/different class(es). For better visualization, only partial negative pairs sampled randomly are shown in this figure.}
\label{fig:distance}
\end{figure}

\begin{table}[]
\caption{The confusion of positive and negative pairs on \emph{MSMT17}. $Confusion$ means the number of overlapped pairs in Figure~\ref{fig:distance}.}
\label{tab:distance}
\centering
\vspace{-0.2cm}
\begin{tabular}{ccc}
\hline
 & $Confusion \downarrow$ & mAP/R1 (\%) $\uparrow$ \\ \hline
Baseline        & 26,469         & 66.1/84.6  \\
$T_1$         & 25,738        & 67.9/85.3  \\
$T_2$         & 25,064        & 68.1/85.4  \\
$Cat.$        & 24,390       & 68.3/85.5  \\ \hline
\end{tabular}
\end{table}

\textbf{Intra-class and inter-class distance.} DC-Former divides the original embedding space into multiple compact subspaces, reducing intra-class distance but also reducing inter-class distance. To verify the effectiveness of the compact space proposed by DC-Former, the intra-class and inter-class distance of DC-Former's embeddings are calculated and visualized in Figure~\ref{fig:distance}. And the confusion of positive and negative pairs in Figure~\ref{fig:distance} is further calculated in Table~\ref{tab:distance}. Compared to baseline, the embedding space of each token in DC-Former is smaller, as is their concatenation. Moreover, the embedding space of $T_2$ is more compact than that of $T_1$, so $T_2$ achieves higher performance. And the confusion of DC-Former is less than baseline, which means that compact embedding space pushes the embeddings of the same class more tightly than the embeddings of different classes. 

\begin{figure}[]
\centering
\subfigure[Impact of $\lambda$]{
\includegraphics[width=3.5cm]{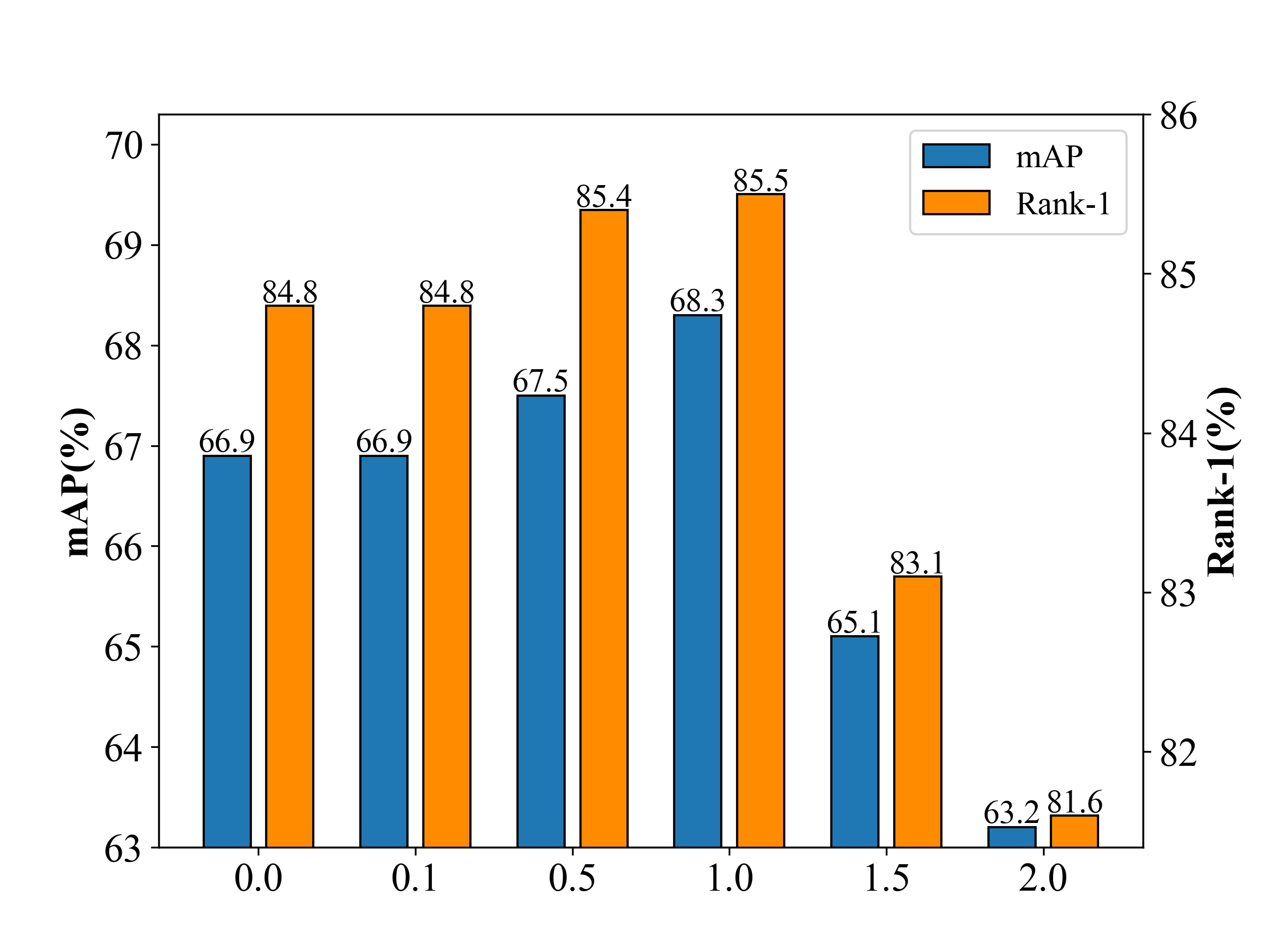}
}
\hspace{-0.4cm}
\subfigure[Impact of $N$]{
\includegraphics[width=4.5cm]{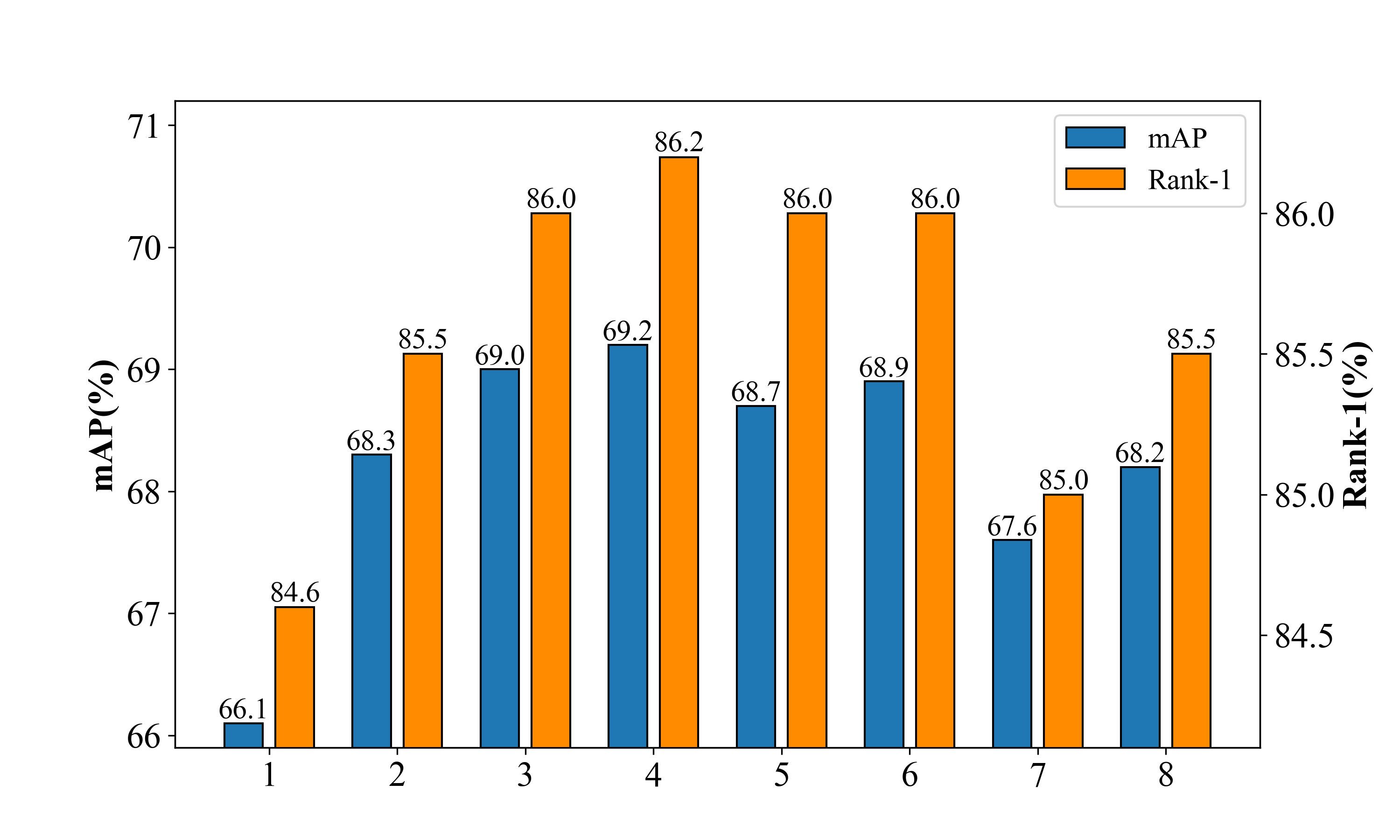}
}
\caption{Visualization of ablation studies on \emph{MSMT17}. Impact of two hyper-parameters of SDC.}
\label{fig:ablation}
\end{figure}

\begin{figure}[!t]
\centering
\vspace{-0.2cm}
\subfigure[w/o DWC]{
\includegraphics[width=4cm]{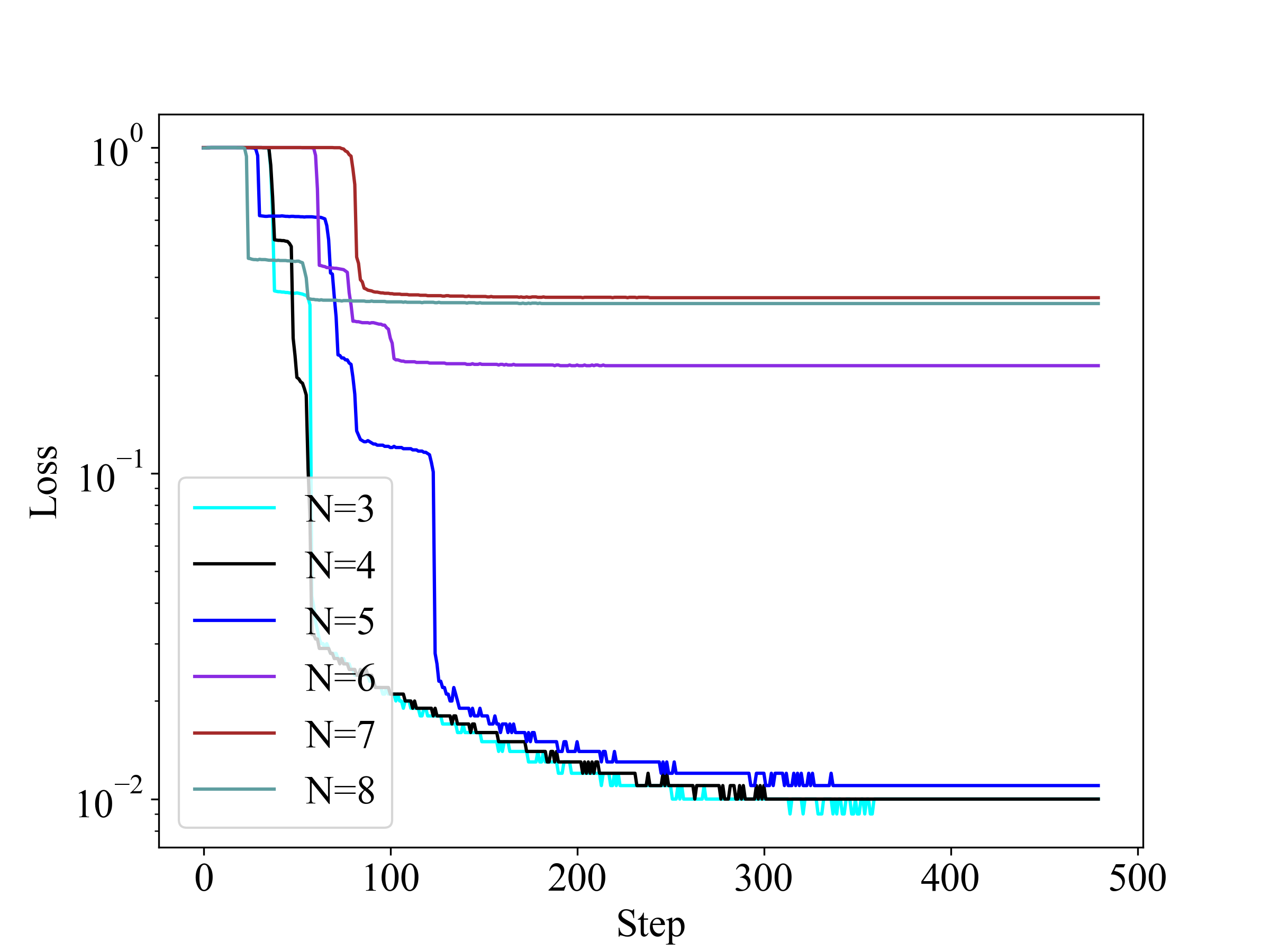}
\label{fig:loss_w/o_dwc}
}
\vspace{-0.2cm}
\hspace{-0.4cm}
\subfigure[with DWC]{
\vspace{-0.2cm}
\includegraphics[width=4cm]{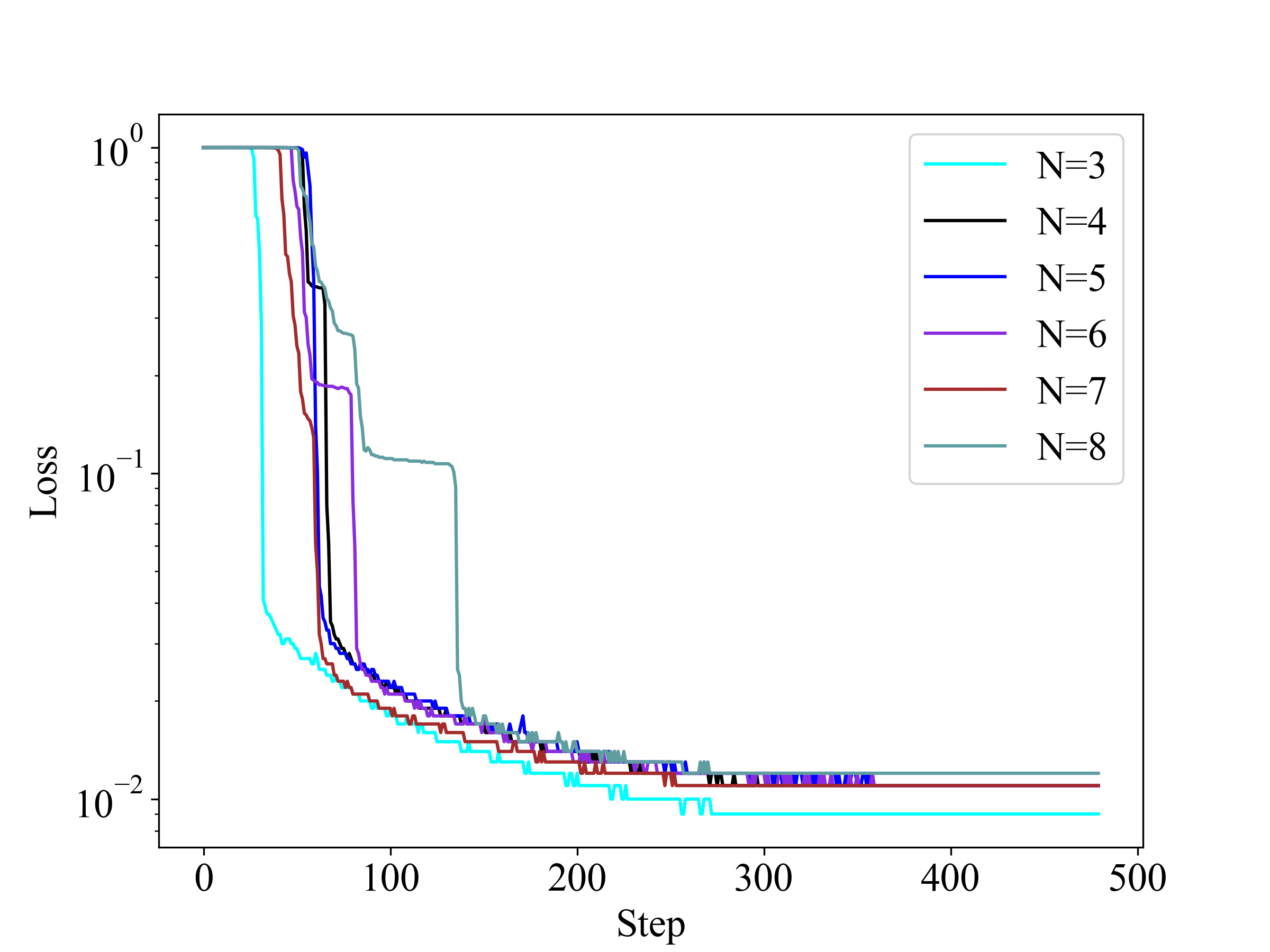}
}
\subfigure[mAP]{
\includegraphics[width=4cm]{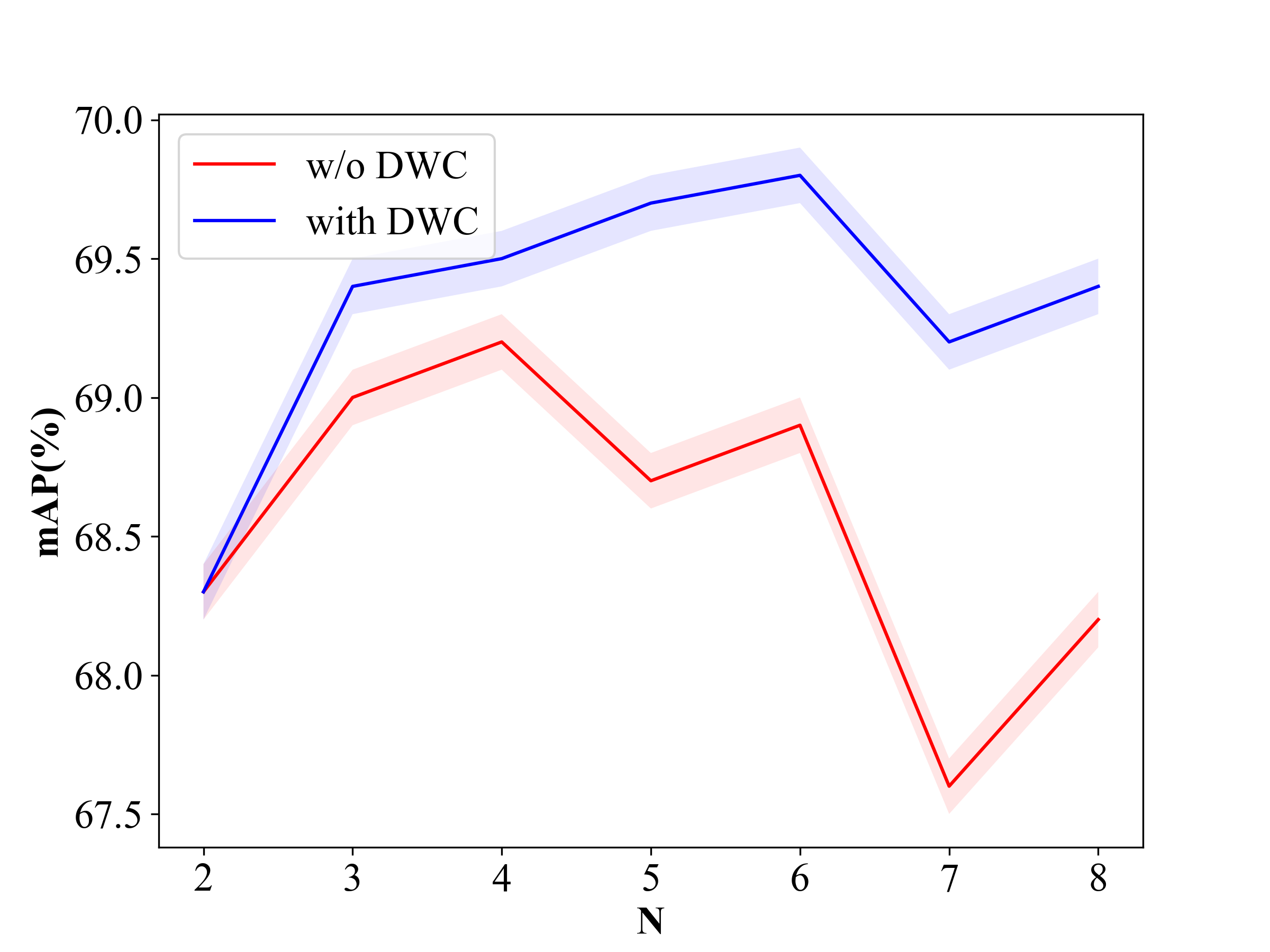}
}
\vspace{-0.2cm}
\hspace{-0.4cm}
\subfigure[Rank-1]{
\includegraphics[width=4cm]{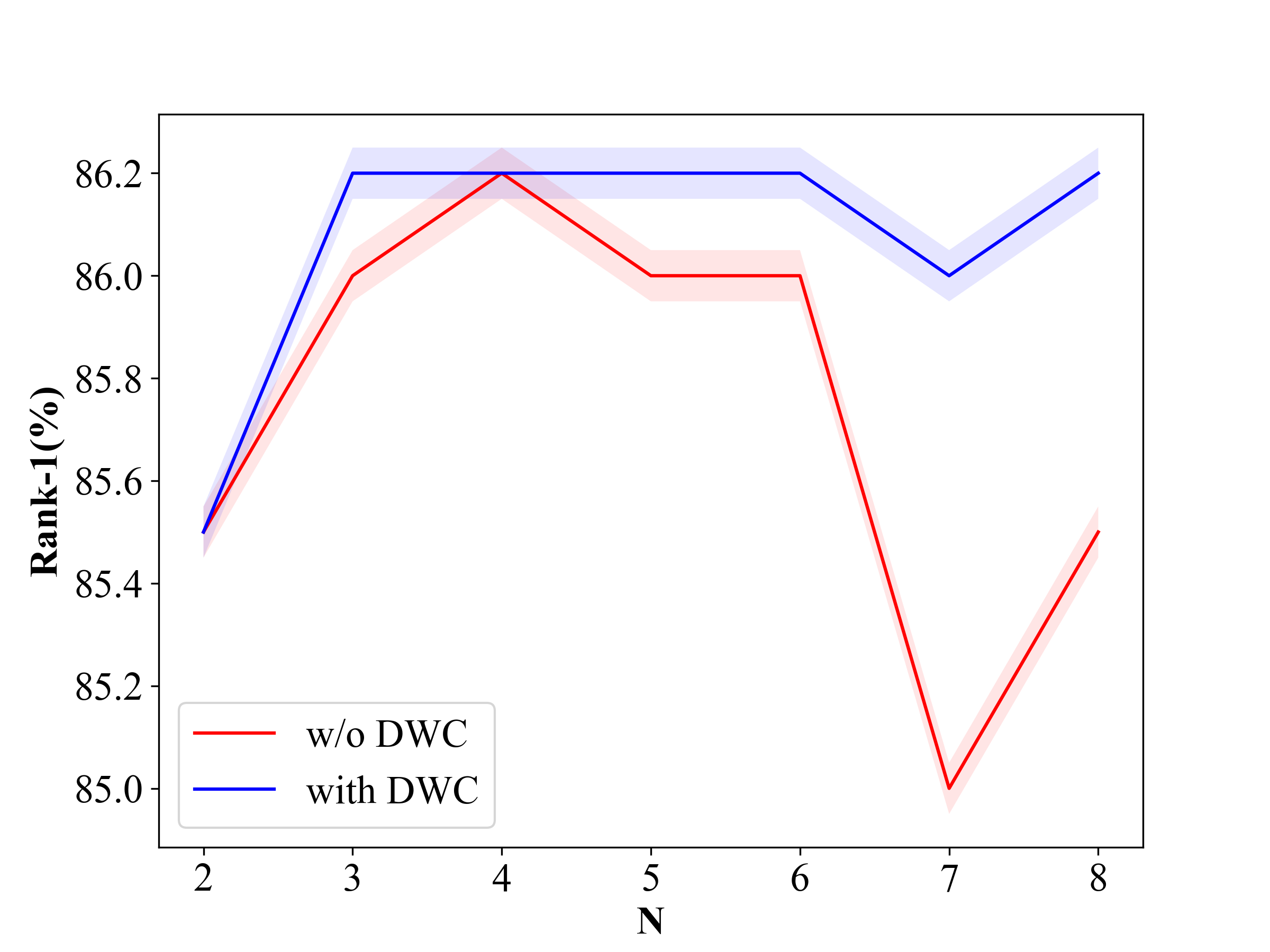}
}
\caption{The evaluation of DWC on \emph{MSMT17}. $N$ denotes the number of class tokens. (a-b) DC-Former's SDC loss when training with/without DWC. (c-d) The mAP and Rank-1 of DC-Former with/without DWC.}
\label{fig:dwc_ablation}
\end{figure}

\textbf{Hyper-parameters of SDC.} SDC has two hyper-parameters which are the weight of loss $\lambda$ and the number of class tokens $N$. We analyze the influence of $\lambda$ on the performance in Figure~\ref{fig:ablation}(a). When $\lambda$ = 0, baseline achieves 66.9\% mAP and 84.8\% Rank-1 on \emph{MSMT17}. When $\lambda$ = 0.1, SDC doesn't work because it's too small. As $\lambda$ increases, the performance increases. When $\lambda$ = 1, the mAP and Rank-1 are improved to 68.3\% and 85.5\%, respectively. When continuing to increase $\lambda$, the performance is degraded because excessive weights make the model pull apart two features at the beginning, making it difficult to optimize the classification loss. Therefore, $\lambda$ = 1 is the best beneficial for learning multiple diverse features. 

The experiments on the number of class tokens $N$ is in Figure~\ref{fig:ablation}(b). Increasing the number of class tokens improves the performance of model. When $N$ is between 3 and 6, mAP and Rank-1 are higher. And $N$ = 4 reaches the best performance at 69.2\% mAP and 86.2\% Rank-1. Continuing to increase $N$, performance is degraded because finding too much embedding space makes training difficult. As we can see SDC loss in Figure~\ref{fig:loss_w/o_dwc}, too much tokens cannot optimize model well because the subspaces cannot be separated due to competition between tokens. $N$ could be a little different in different datasets. In the SOTA experimental configuration (Table~\ref{tab:sota_person}), $N$ is [6,5,4] for \emph{MSMT17}, \emph{Market-1501} and \emph{CUHK03}, respectively.

\textbf{Dynamic Weight Controller.} The effectiveness of the proposed DWC module is validated in Figure~\ref{fig:dwc_ablation}. The SDC loss in training phase illustrated in Figure~\ref{fig:dwc_ablation}(a-b) shows that DWC can make each pairs learn evenly so that they are all orthogonal to each other, which shows effectiveness on balancing the relative importance among class tokens during training as the token number increases. The performance in Figure~\ref{fig:dwc_ablation}(c-d) shows that DWC has limited effect when $N$ is small (less than 5). While it provides about +1.0\% improvement when $N$ is large. When $N\geq7$, the performance decreases because the number of subspaces in a finite embedding space has reached its limit. Too small space makes the embeddings lose their discrimination.

\begin{figure}[]
\centering
\subfigure[mAP]{
\vspace{-0.2cm}
\includegraphics[width=4cm]{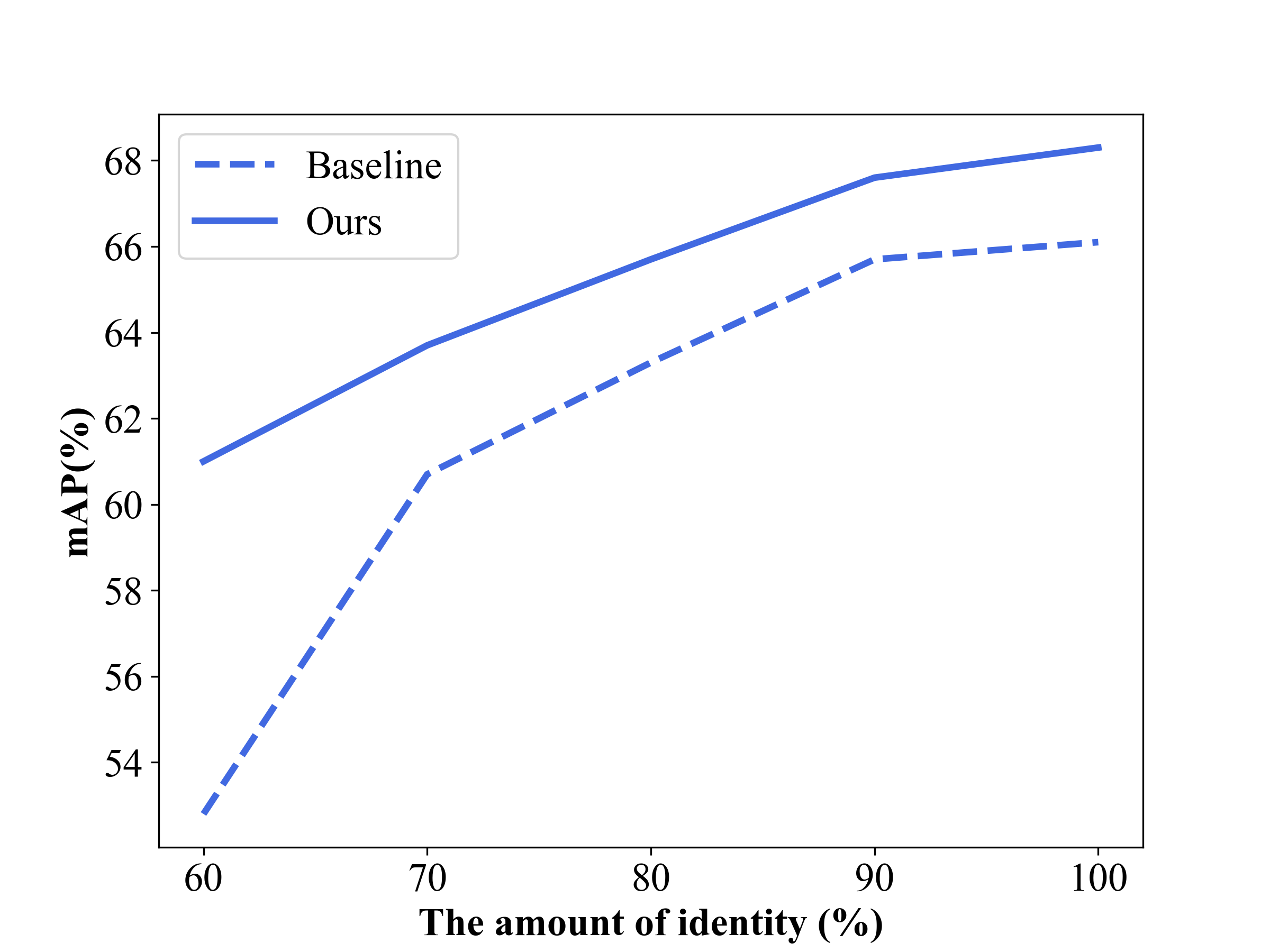}
}
\vspace{-0.2cm}
\hspace{-0.4cm}
\subfigure[Rank-1]{
\includegraphics[width=4cm]{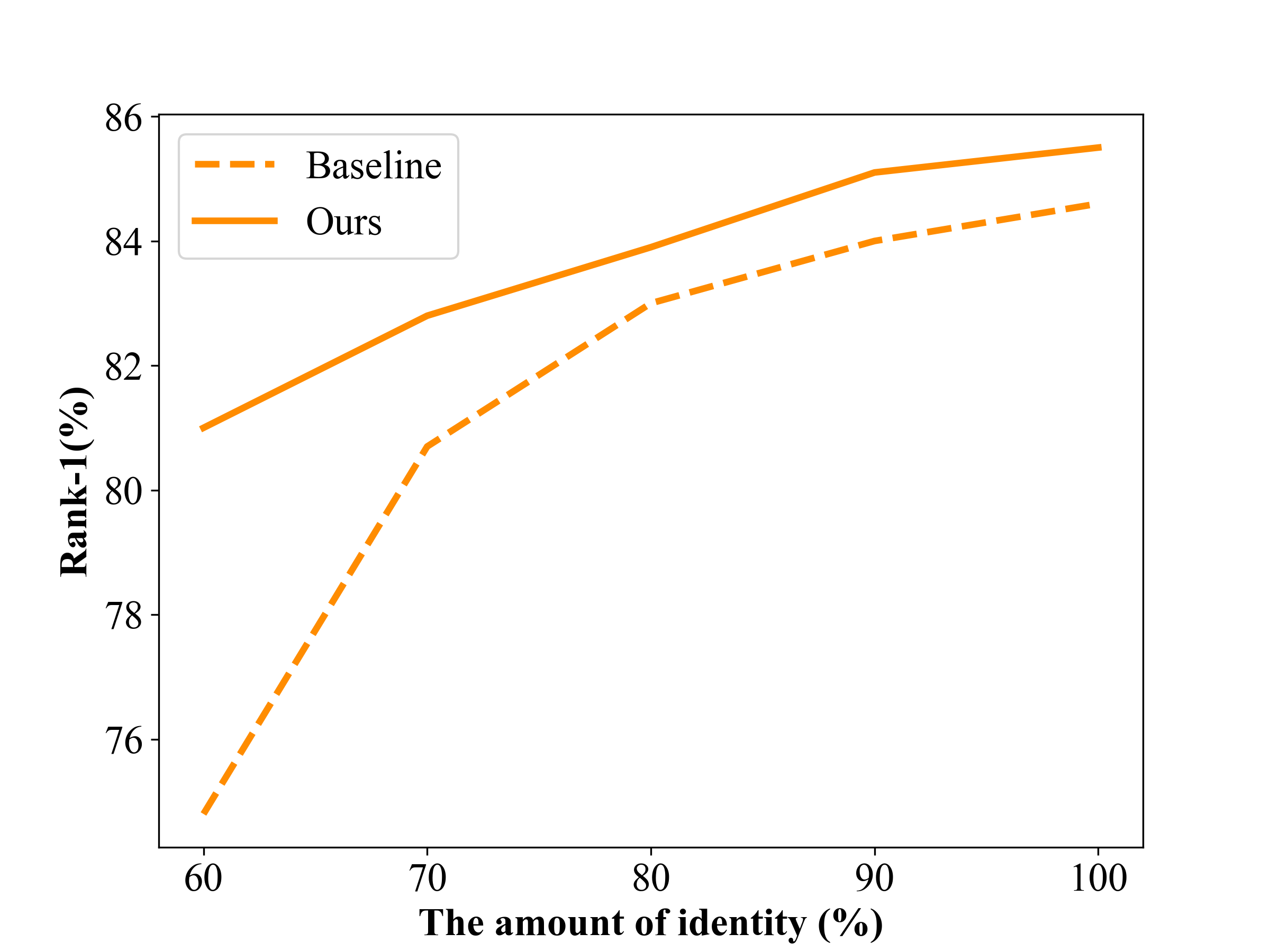}
}
\caption{Performance under training with a smaller amount of identity on \emph{MSMT17}. Randomly select partial categories as training set, and keep the test set unchanged.}
\label{fig:amount_data}
\end{figure}
\textbf{Effectiveness with a smaller amount of identity.} We evaluate the effectiveness of DC-Former with a smaller amount of identity in Figure~\ref{fig:amount_data}. DC-Former achieves comparable results by using less than 20\% identities of baseline. And the smaller the amount of identity, the more obvious the advantages of DC-Former. Increasing the amount of identity strengthens the ability of the model to identify similar identities. And more compact embedding space of DC-Former also enhances the discrimination of similar classes' representations, which is similar to the effect of increasing the amount of identity.

\section{Conclusions}
In this paper, we propose a transformer-based network for re-ID to learn multiple diverse and compact embedding subspaces, which improves the robustness of representation by increasing identity density of embedding space. And the fusion of these representations from different subspaces further improves performance. Our method outperforms previous state-of-the-arts on three person re-ID benchmarks. Based on this promising results, we believe our method has great potential to be further explored in other area, and we hope it can bring new insights to the community.

\clearpage

\bibliography{aaai23}

\end{document}